\documentclass{article}

\usepackage{arxiv}

\usepackage[utf8]{inputenc} 
\usepackage[T1]{fontenc}    
\usepackage{hyperref}       
\usepackage{url}            
\usepackage{booktabs}       
\usepackage{amsfonts}       
\usepackage{nicefrac}       
\usepackage{microtype}      
\usepackage{lipsum}		
\usepackage{graphicx}
\usepackage{natbib}
\usepackage{doi}
\usepackage{multirow}
\usepackage{mathtools}
\usepackage{textcomp}

\title{DocSCAN: Unsupervised Text Classification via Learning from Neighbors}

\author{ Dominik Stammbach \\
  ETH Zurich \\
  \texttt{dominsta@ethz.ch} \\\And
	\And
    Elliott Ash \\
  ETH Zurich \\
  \texttt{ashe@ethz.ch} \\}

\renewcommand{\shorttitle}{DocSCAN}

\hypersetup{
pdftitle={A template for the arxiv style},
pdfsubject={q-bio.NC, q-bio.QM},
pdfauthor={David S.~Hippocampus, Elias D.~Striatum},
pdfkeywords={Text Classification, Unsupervised Learning, Neighbor-based Clustering},
}

\begin{document}
\maketitle

\begin{abstract}
We introduce DocSCAN, a completely unsupervised text classification approach built on the \textit{Semantic Clustering by Adopting Nearest-Neighbors} algorithm. For each document, we obtain semantically informative vectors from a large pre-trained language model. We find that similar documents have proximate vectors, so neighbors in the representation space tend to share topic labels. Our learnable clustering approach then uses pairs of neighboring datapoints as a weak learning signal to automatically learn topic assignments. On three different text classification benchmarks, we improve on various unsupervised baselines by a large margin.
\end{abstract}


\section{Introduction}

"What is this about?" is the starting question in human and machine reading of text documents. While this question would invite a variety of answers for documents in general, there is a large set of corpora for which each document can be labeled as belonging to a singular category or topic. Text classification is the task of automatically mapping texts into these categories. In the standard supervised setting \cite{Vapnik2000}, machine learning algorithms learn such a mapping from annotated examples. Annotating data is costly, however, and the resulting annotations are usually domain-specific. Unsupervised methods promise to reduce the number of labeled examples needed or to dispense with them altogether. 

This paper builds on recent developments in the domain of unsupervised neighbor-based clustering of images, the SCAN algorithm: \textit{Semantic Clustering by Adopting Nearest neighbors} \cite{scan}. We adapt the algorithm to text classification and report strong experimental results on three text classification benchmarks. The intuition behind SCAN is that images often share the same label, if their embeddings in some representation space are close to each other. Thus, we can leverage this regularity as a weakly supervised signal for training models. We encode a datapoint and its neighbors through a network where the output of the network is determined by a classification layer. The model learns that it should assign similar output probabilities to a datapoint and each of its neighbors. In the ideal case, model output is consistent and one-hot, i.e. the model confidently assigns the same label to two neighboring datapoints. 

Deep Transformer networks have led to rapid improvements in text classification and other natural language processing (NLP) tasks \citep[see e.g.][]{xlnet}. We draw from such models to obtain task-agnostic contextualized language representations. We use SBERT embeddings \cite{sbert}, which have proven performance in a variety of downstream tasks, such as retrieving semantically similar documents and text clustering. We show that in this semantic space, indeed neighboring documents tend to often share the same class label and we can use this proximity to build a dataset on which we apply our neighbor-based clustering objective. We find that training a model exploiting this regularity works well for text classification and outperforms a standard unsupervised baseline by a large margin. All code for DocSCAN can be found publicly available online.\footnote{\url{https://github.com/dominiksinsaarland/DocSCAN}}

\section{Related Work}

Unsupervised learning methods are ubiquitous in natural language processing and text classification. For a more general overview, we refer to surveys discussing the topic in extensive details \citep[see e.g.][]{text_mining_handbook,text_as_data,text_clustering_review, text_classification_techniques_review,li2021survey_text_classification}. One common approach for text classification is to represent documents as vectors and then apply any clustering algorithm on the vectors \cite{text_clustering_review,brief_survey_text_clustering}. The resulting clusters can be interpreted as the text classification results. A popular choice is to use the k-means algorithm which learns cluster centroids that minimize the within-cluster sum of squared distances-to-centroids \citep[see e.g.][]{kmeans_2,kmeans_4,kmeans_5,kmeans_3,kmeans_1,kmeans_6}. This methodology has also applications in social science research, where for example \citet{demszky2019analyzing} classify tweets using this method. K-means can also be applied in an iterative manner \cite{iterative_clustering}. 

There exist more sophisticated methods for generating weak labels for unsupervised learning for text classification. However, most of these methods take into account some sort of domain knowledge or heuristically generated labels. For example, \citet{snorkel} generate a correlation-based aggregate of different labeling functions to generate proxy labels. \citet{cosine_weak_supervision} create weak labels via heuristics, and \citet{meng-etal-2020-text} use seed words (most importantly the label name) and infer the text category assignment from a masked language modeling task and seed word overlap for each category. DocSCAN is not subject to any of these dependencies. Similarly to k-means, we only need the number of topics present in a dataset. Hence, we think it is well suited to be compared against k-means.


\section{Method}
\label{sec:method}

In this work, we build on the SCAN algorithm \cite{scan}. It is based on the intuition that a datapoint and its nearest neighbors in (some reasonable) representation space often share the same class label. The algorithm consists of three stages: (1) learn representations via a self-learning task, (2) mine nearest neighbors and fine-tune a network on the weak signal that two neighbors share the same label, and (3) confidence-based self-labeling of the training data (which is ommitted in this work\footnote{The authors use heavily augmented images for the confidence-based self-labeling step. There is no straightforward translation of this approach to NLP. Tokens are discrete, symbolic characters, rather than the continuous quantities contained in pixels. We skip this step and leave exploration to future work.}).

\begin{figure}
    \centering
    \includegraphics[width=0.5\textwidth]{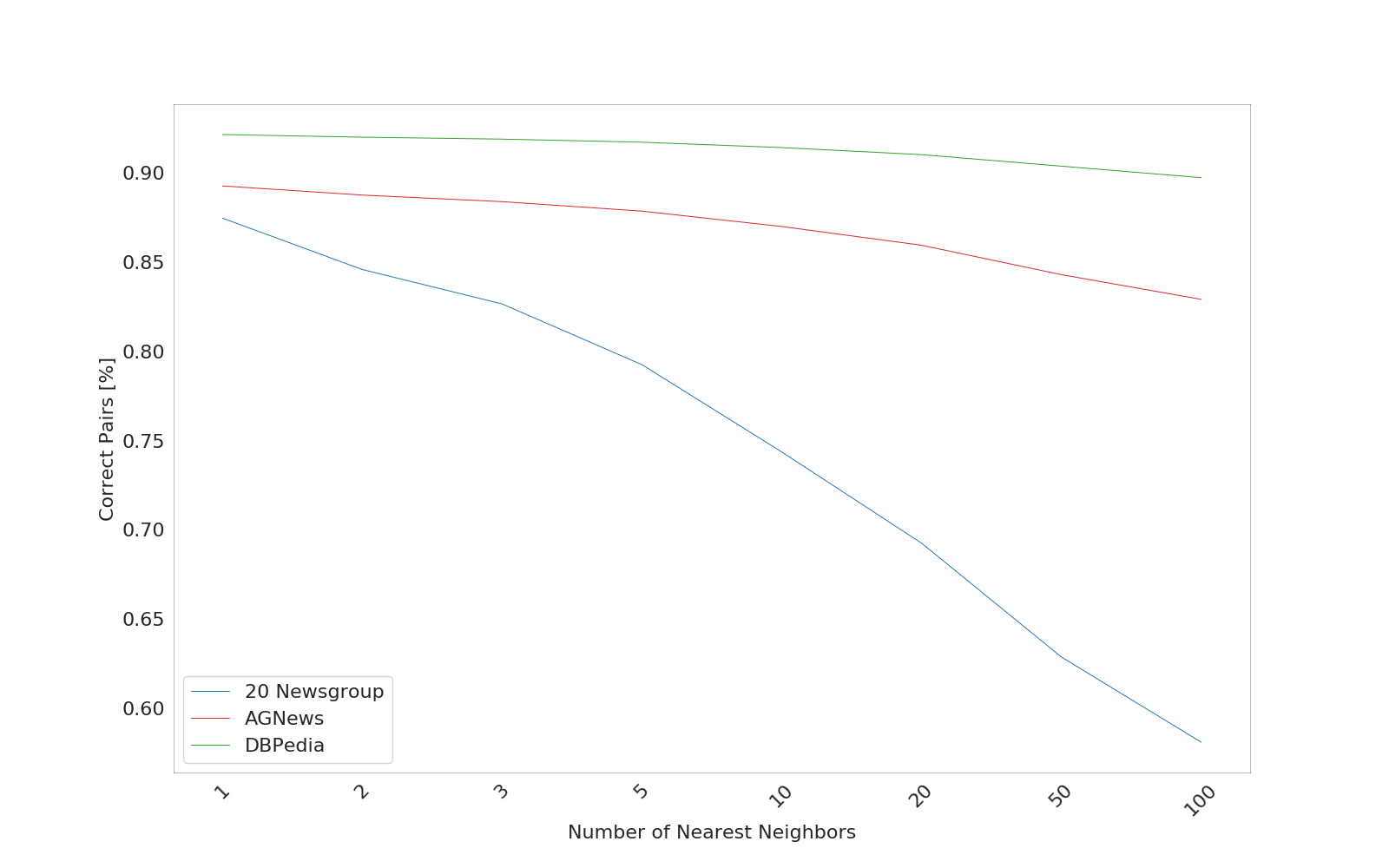}
    \caption{Accuracy of datapoint/neighbor pairs sharing the same label for different text classification benchmarks.}
    \label{fig:label_accuracy_text}
\end{figure}

Our adaptation DocSCAN to text classification works as follows. In Step 1, we need a document embedding method that serves as an analogue to SCAN's self-learning task for images. Textual Entailment \cite{nli} is an interesting pre-training task yielding transferable knowledge and generic language representations, as already shown in \cite{infersent}. Combining this pre-training task and large Transformer models, e.g., \cite{devlin2019bert} has led to SBERT \cite{sbert}: A network of BERT models fine-tuned on the Stanford Natural Language Inference corpus \cite{snli}. SBERT yields embeddings for short documents with proven performance across domains and for a variety of tasks, such as semantic search and clustering. For a given corpus, we apply SBERT and get a 768-dimensional dense vector for each document.\footnote{We also experimented with other document representations. We discuss results in more detail in Appendix \ref{app:ablation}} We directly use the pre-trained SBERT model fine-tuned on top of the MPNet model\footnote{The \textit{all-mpnet-base-v2} model taken from \url{https://www.sbert.net/docs/pretrained\_models.html}} \cite{song2020mpnet}, which yields the best\footnote{"best" embeddings at the time of submission of this work} (on average) performing embeddings for 14 sentence embedding tasks and 6 semantic search tasks. 

Step 2 is the mining of neighbors in the embedding space. We apply Faiss \cite{faiss} to get Euclidean distances between all embedded document vectors. The retrieved neighbors are the documents having the smallest Euclidean distance to a reference datapoint.

SCAN worked because images with proximate embeddings tended to share class labels. Is that the case with text? Figure \ref{fig:label_accuracy_text} shows that the answer is yes: across three text classification benchmarks, neighboring document pairs do indeed often share the same label. The fraction of pairs sharing the same label at $k=1$ is above 85\% for all datasets examined. For $k=5$, the resulting fraction of correct pairs (from all mined pairs) is still higher than 75\% in all cases. Furthermore, these frequencies of correct pairs for $k=5$ are often higher than the frequency of correct pairs reported for images in \cite{scan}.

Next, we describe the SCAN loss,

\begin{equation}
\begin{multlined}
     -\frac{1}{|\mathcal{D}|} \sum_{x \in \mathcal{D}} \sum_{k \in \mathcal{N}_x} log (f(x) \cdot f(k)) + \lambda \sum_{i \in \mathcal{C}} p_i log (p_i)
    \label{eq:scan_loss}
\end{multlined}
\end{equation}

which can be broken down as follows. The first part of Eq. (\ref{eq:scan_loss}) is the consistency loss. Our model $f$ (parametrized by a neural net) computes a label for a datapoint $x$ from the dataset $\mathcal{D}$ and for each datapoint $k$ in the set of the mined neighbors from $x$ in $\mathcal{N}_x$. We then simply compute the dot product (denoted as $\cdot$) between the output distribution (normalized by a softmax function) for our datapoint $x$ and its neighbor $k$. This dot product is maximized if both model outputs are one-hot with all probability mass on the same entry in the respective vectors. It is consistent because we want to assign the same label for a datapoint and all its neighbors. The second term is an auxiliary loss to obtain regularization via entropy (scaled by a weight $\lambda$), such that the model is encouraged to spread probability mass across all clusters $\mathcal{C}$ where $p_i$ denotes the assigned probability of cluster $i$ in $\mathcal{C}$ by the model. Without this entropy term, there exists a shortcut by collapsing all examples into one single cluster. The entropy term ensures that the distribution of class labels resulting from applying DocSCAN tends to be roughly uniform. Thus, it works best for text classification tasks where the number of examples per class is balanced as well.

To summarize: We use SBERT and embed every datapoint in a given text classification dataset. We then mine the five nearest neighbors for every datapoint. This yields our weakly supervised training set. We fine-tune networks on neighboring datapoints using the SCAN loss. At test time, we compute $f(x)$ for every datapoint $x$ in the test set. We set the number of outcome classes equal to the numbers of classes in our considered datasets and use the hungarian matching algorithm \cite{Kuhn55thehungarian} to obtain the optimal cluster-to-label assignment. 
\section{Experiments}\label{sec:experiments}

We apply DocSCAN on three widely used but diverse text classification benchmarks: The 20NewsGroup data \cite{Newsgroups20}, the AG's news corpus \cite{ag_news}, and lastly the DBPedia ontology dataset \cite{dbpedia}. We provide further dataset descriptions in Appendix Section \ref{app:dataset}. 

\begin{table*}[t!]
\footnotesize
\centering
\renewcommand{\arraystretch}{1.5}
\begin{tabular}{p{5.2cm} p{2.1cm} p{2.1cm} p{2.1cm}}
 \hline   
Experiment & 20 News & AG news & DBPedia \\ \hline
[1] Random Baseline & 7.0 \textpm 0.0 & 26.1 \textpm 0.3 & 7.7 \textpm 0.0 \\
{[2]} TF-IDF + k-means &  32.6 \textpm 1.1 & 49.5 \textpm 6.3 & 47.6 \textpm 3.0 \\
{[3]} SBERT embeddings + k-means & 54.2 \textpm 1.6 & 69.2 \textpm 7.3 & 76.9 \textpm 4.3 \\ \hline
{[4]} DocSCAN & \textbf{59.4} \textpm 1.9 & \textbf{84.1} \textpm 2.6 & \textbf{84.6} \textpm 3.8 \\ \hline
{[5]} SBERT embeddings + SVM & 82.7 & 92.1 & 98.7  \\ \hline
{6]} Related Literature & 58.2 & 84.52 \textpm 0.50 & 91.1 \\ \hline

\end{tabular}
\caption{Test-set accuracy by benchmark dataset (columns) and classifier (rows). Cell values give the mean over 10 runs with 95\% confidence interval. Note that the results reported from the related literature in the last row might not be directly comparable to our method due to different experimental setups. The 20 News results are taken from \cite{chu-etal-2021-unsupervised}, the AG news results from \cite{iterative_clustering}, and the DBPedia results from \cite{meng-etal-2020-text}. }
  \label{tab:results_topic_classification}
\end{table*}

The main results are reported in Table \ref{tab:results_topic_classification}. For all experiments, we report the mean accuracy over 10 runs on the test set (with different seeds and the 95\% confidence interval). The columns correspond to the benchmark corpora. The rows correspond to the models, starting with a random baseline [1], two k-means baselines [2, 3] and the results obtained by DocSCAN in [4]. We also report a supervised learning baseline [5] and results taken from related literature in [6].

Row [1] provides a sensible lower-bound, row [5] analogously a supervised upper-bound for text classification performance. In the random draw [1], accuracy by construction converges to the average of the class proportions. The supervised model [5] is an SVM classifier applied to the same SBERT embeddings\footnote{We also trained the SVM classifier with TF-IDF representations and obtained similar results for all experiments.} which serve as inputs to the k-means baseline and to DocSCAN. Predictably, the supervised baseline obtains strong accuracy on these benchmark classification tasks.

The industry workhorse for clustering is k-means, an algorithm for learning cluster centroids that minimize the within-cluster sum of squared distances-to-centroids. When applied to TF-IDF-weighted bag-of-n-grams features [2], k-means improves over the results obtained in [1]. When applied to SBERT vectors [3], we see large improvements over all previous experiments. These results suggest that k-means applied to reasonable document embeddings already yields satisfactory results for text classification. Second, they corroborate what we already saw in Figure \ref{fig:label_accuracy_text}, that neighbors in SBERT representation space contain information about text topic classes. 

So what does DocSCAN add? We fine-tune a classification layer using the SBERT embeddings with the SCAN objective (Eq. \ref{eq:scan_loss}) and $k=5$ neighbors. We observe unambiguous and significant improvements over the already strong k-means baseline in all three datasets (as we can judge from the 95\% confidence intervals). The smallest improvements (over 5\% points) are made on the 20 News dataset, containing 20 classes. The largest improvement gains are observed for AG news with 4 classes, suggesting that DocSCAN above all works best for text classification tasks with a lower number of classes. Surprisingly, we do not find that the improvements correlate with the accuracy of neighboring pairs sharing the same label (see Figure \ref{fig:label_accuracy_text}), but rather with the numbers of classes in the dataset (see Table \ref{tab:dataset_statistics}). In the case of the AG news data with only a few different classes, we find that DocSCAN approaches the performance of a supervised baseline using the same input features. 

Finally, in [6] we show results from related literature on unsupervised text classification. We find that DocSCAN performs comparable to other completely unsupervised methods. We find that DocSCAN obtains the best results for the 20 News dataset, comparable results in the case of AG news data and slightly worse results than the related literature on the DBPedia data. However, we note that DocSCAN is a simple method consisting of only $hidden\_dim * num\_classes$ parameters, that is exactly one classification layer which is fine-tuned in a completely unsupervised manner using the SCAN loss. Whereas the results for DBPedia from \cite{meng-etal-2020-text} are obtained by fine-tuning whole language models using domain knowledge (seed words). 

We show and discuss ablation experiments for DocSCAN in Appendix \ref{app:ablation}. Specifically, we conduct experiments regarding the various hyper-parameters of the algorithm and find that it is robust to such choices. Furthermore, we find that DocSCAN outperforms a k-means baseline over different input features in all settings. Given the findings derived from these experiments, we recommend default hyperparameters for applying DocSCAN.

\section{Conclusion}

In this work, we introduced DocSCAN for unsupervised text classification. Analogous to the recognizable object content of images, we find that a document and its close neighbors in embedding space often share the same class in terms of the topical content. We show that this consistency can be used as a weak signal for fine-tuning text classifier models in an unsupervised fashion. We start with SBERT embeddings and fine-tune DocSCAN on three text classification benchmarks. We outperform a random baseline and two k-means baselines by a large margin. We discuss the influence of hyper-parameters and input features for DocSCAN and recommend default parameters which we have observed to work well across our main results. 

As with images, unsupervised learning with SCAN can be used for text classification. However, the method may not work as generically, and should for example be limited to text classification in cases of balanced datasets (given that we use an entropy loss as an auxiliary objective). Still, this work points to the promise of further exploration of unsupervised methods using embedding geometry.

\bibliographystyle{unsrtnat}

\clearpage

\bibliography{references_new}  


\clearpage
\appendix
\onecolumn

\section{Dataset Statistics}
\label{app:dataset}

\begin{table}[!ht]
    \footnotesize
    \centering
    \begin{tabular}{l c c c  p{8cm}}
         Dataset & \# Examples  & \# Classes  & Avg. Length & Example \\ \hline
        20News & 11'314 & 20 & 248 & [...] I Have a Sound Blaster ver 1.5 When I try to install driver ver 1.5 (driver that comes with window 3.1) [...] \\ 
        AG's Corpus & 120'000 & 4 & 31 & Wall St. Bears Claw Back Into the Black (Reuters) Reuters - Short-sellers, Wall Street's dwindling band of ultra-cynics, are seeing green again. \\
        DBPedia & 560'000 & 14  & 46 & Abbott of Farnham E D Abbott Limited was a British coachbuilding business based in Farnham Surrey trading under that name from 1929. A major part of their output was under sub-contract to motor vehicle manufacturers. Their business closed in 1972. \\  \hline
    \end{tabular}
            \caption{Dataset Statistics}
    \label{tab:dataset_statistics}
\end{table}

We apply DocSCAN to three diverse datasets widely used in unsupervised text classification: (1) The 20NewsGroup data contains text from UseNet discussion groups (20 classes). (2) The AG's news corpus \cite{ag_news}, which consists of the title and description field of news articles (4 classes). And lastly the DBPedia ontology dataset \cite{dbpedia} which includes titles and abstracts of Wikipedia articles (14 classes). 

In Table \ref{tab:dataset_statistics}, we show the numbers of training examples, number of classes,  the average document length and one text example from each dataset. We selected these datasets because they are established standard datasets for unsupervised text classification. The three datasets vary in domain, number of classes, and text lengths. But they have all in common that the number of examples per class are roughly balanced, hence DocSCAN is well suited to tackle these datasets. 

\section{Ablation Experiments}
\label{app:ablation}

In Table \ref{tab:ablation_experiments}, we report how DocSCAN performs under various different hyper-parameters which possibly could affect the performance of the algorithm. In the two last columns, we report the mean accuracy of 10 runs (and the 95\% confidence interval) on the AG news and DBPedia training datasets. As common practice in unsupervised learning, we cluster the dataset and then report evaluation metrics on the training set itself (whereas in the main results, we discuss the performance on the test sets of the respective datasets). 

We investigate the number of neighbors considered (A), the weight of the entropy loss (B), batch sizes (C), dropout (D) and number of epochs (E). To optimize the SCAN loss, we use Adam \cite{adam} with default parameters in all experiments. DocSCAN runs somewhat stable across different choices of these hyperparameters, yielding similar results which all outperform the k-means baseline by a large margin. The two worst performances are achieved if we either set the entropy weight too low ($\lambda=1$) or do not consider enough neighboring pairs ($k=2$). The influence of all other hyperparameters seems limited. We recommend using the default parameters reported in the first row. The main results in Table \ref{tab:results_topic_classification} were obtained using this set of hyperparameters.

\begin{table}[!ht]
\footnotesize
\centering
\renewcommand{\arraystretch}{1.5}
\begin{tabular}{c | c c c c c| c c}
 \hline   
& Neighbors & Entropy Weight & Batch Size & Dropout & Epochs & Accuracy AG news & Accuracy DBPedia  \\ \hline
DocSCAN & 5 & 2 & 128 & 0.1 & 5 & 83.2 \textpm 3.8 & 85.8 \textpm 3.5 \\ \hline
\multirow{3}{*}{(A)} & 2 & & & & & 77.5 \textpm 6.7 & 83.1 \textpm 5.1 \\
 & 3 & & & & & 78.4 \textpm 5.5 & 85.3 \textpm 3.0 \\ 
 & 10 & & & & & 82.4 \textpm 5.6 & 86.1 \textpm 3.5 \\ \hline
\multirow{2}{*}{(B)} & & 1 & & & & 75.8 \textpm 5.3 & 80.3 \textpm 2.8 \\
 & & 4 & & & & 80.4 \textpm 3.5 & 86.7 \textpm 2.8 \\ \hline
\multirow{2}{*}{(C)} & &  & 64 & & & 82.4 \textpm 5.0 & 87.5 \textpm 4.2 \\
 & &  & 256 & & & 81.3 \textpm 4.3 & 84.6 \textpm 4.1 \\ \hline
\multirow{2}{*}{(D)} & &  &  & 0 & &  81.9 \textpm 3.7 & 86.1 \textpm 4.4 \\
 & &  &  & 0.33 & & 80.3 \textpm 3.9 & 86.8 \textpm 2.6 \\ \hline
\multirow{2}{*}{(E)} & &  &  &  & 3 & 79.4 \textpm 5.3 & 84.2 \textpm 4.4 \\
 & &  &  &  & 10 & 81.5 \textpm 3.4 & 84.7 \textpm 3.7 \\ \hline
k-means &  & & & & & 66.2 \textpm 8.2 & 77.1 \textpm 4.9 \\ \hline
\end{tabular}
\caption{Ablation Studies for DocSCAN Hyper-parameters (results reported on the AG news and DBPedia training set, cell values give the mean over 10 runs with 95\% confidence interval).}
\label{tab:ablation_experiments}
\end{table}


We also investigate whether the success of DocSCAN for text classification stems from the chosen document embeddings. For this, we consider a number of different input features for the algorithm and run DocSCAN with these features, holding everything else constant. We show results in Table \ref{tab:ablation_experiments_input_features}. We report the mean performance of 10 runs and the 95\% confidence interval on the AG news training set.

We run several document embedding techniques, starting with the TF-IDF-weighted bag-of-n-grams \cite{tfidf}. Second, we consider the averaged GloVe embeddings of all words in a document \cite{pennington-etal-2014-glove}, Universal Sentence Encoder (USE) embeddings \cite{universal_sentence_encoder} and lastly the performance of DocSCAN using SBERT embeddings \cite{sbert}. We observe again that DocSCAN performs better than k-means in every setting. However, the performance gap for different features varies. For example, we observe best k-means performance using USE embeddings, whereas the best DocSCAN performance is achieved via SBERT embeddings. Also, TF-IDF + k-means yields rather mixed results, whereas TF-IDF + DocSCAN performs more than 20\% points better. In light of these results, we recommend to use SBERT embeddings if considering applying DocSCAN to other work.

\begin{table}[!ht]
\footnotesize
\centering
\renewcommand{\arraystretch}{1.5}

\begin{tabular}{p{3cm}  p{3cm} p{3cm} }
 \hline   
Features & k-means & DocSCAN \\ \hline
TF-IDF & 53.9 \textpm 4.1 & 76.8 \textpm 4.3 \\ 
avg. GloVe & 55.4 \textpm 3.6 & 59.3 \textpm 0.3 \\
USE Embeddings & 74.4 \textpm 8.3 & 79.1 \textpm 8.6 \\
SBERT & 66.2 \textpm 8.2 & 83.2 \textpm 3.8 \\ \hline
\end{tabular}
\caption{Ablation Studies for Different Input Features (results reported on the AG news training set, cell values give the mean over 10 runs with 95\% confidence interval).}
\label{tab:ablation_experiments_input_features}
\end{table}

\end{document}